\pdfoutput=1

\documentclass[11pt]{article}

\usepackage[]{ACL2023}

\usepackage{times}
\usepackage{latexsym}
\usepackage{makecell}
\usepackage{amsmath}
\usepackage{times}
\usepackage{latexsym}
\usepackage{graphicx}
\usepackage{booktabs}
\usepackage{colortbl}
\usepackage{tablefootnote}
\usepackage{multirow}
\usepackage{microtype}
\usepackage{amsfonts}
\usepackage{inconsolata}
\usepackage{titlesec}
\makeatletter
\newcommand{\thickhline}{%
    \noalign {\ifnum 0=`}\fi \hrule height 1.1pt
    \futurelet \reserved@a \@xhline
}
\makeatother

\usepackage[T1]{fontenc}

\usepackage[utf8]{inputenc}

\usepackage{microtype}
\usepackage{inconsolata}

\title{
\textsc{UniSumm} and \textsc{SummZoo}: \\Unified Model and Diverse Benchmark for Few-Shot Summarization
}

\author{
  Yulong Chen$^{1, 2}$ \Thanks{~Yulong Chen completed this work during his internship at Microsoft.} \quad
  Yang Liu$^3$  \Thanks{~Yang Liu is the corresponding author.}\quad
  Ruochen Xu$^3$ \quad
  Ziyi Yang$^3$ \quad\\
  {\bf
  Chenguang Zhu$^3$ \quad
  Michael Zeng$^3$ \quad
  Yue Zhang$^{2, 4}$ \quad} \\
  $^1$ Zhejiang University \quad $^2$ Westlake University \quad $^3$ Microsoft Research \\
  $^4$ Westlake Institute for Advanced Study\\
  \emph{\{chenyulong, zhangyue\}@westlake.edu.cn~~~yaliu10@microsoft.com
  }}

\begin{document}
\maketitle
\begin{abstract}
The high annotation costs and diverse demands of various summarization tasks motivate the development of few-shot summarization.
However, despite the emergence of many summarization tasks and datasets, the current training paradigm for few-shot summarization systems ignores potentially shareable knowledge in heterogeneous datasets.
To this end, we propose \textsc{UniSumm}, a unified few-shot summarization model pre-trained with multiple summarization tasks and can be prefix-tuned to excel at any few-shot summarization task.
Meanwhile, to better evaluate few-shot summarizers, under the principles of diversity and robustness, we assemble and release a new benchmark \textsc{SummZoo}. 
It consists of $8$ summarization tasks with multiple sets of few-shot samples for each task, covering diverse domains.
Experimental results and analysis show that \textsc{UniSumm} outperforms strong baselines by a large margin across all sub-tasks in \textsc{SummZoo} under both automatic and human evaluations and achieves comparable results in human evaluation compared with a GPT-3.5 model.
\end{abstract}

\section{Introduction}
There has been a recent surge of interest in summarizers based on large pre-trained language models (PLMs)~\cite{liu-lapata-2019-text,yang-etal-2020-ted, zhong-etal-2020-extractive, yu2022survey, xu-etal-2022-narrate, wang2023element}, where
various summarization tasks (the term \emph{task} later in this paper refers to a specific summarization task, e.g., query-focused meeting summarization, which is usually associated with a corresponding dataset, e.g., QMSum, unless otherwise specified.) have been proposed to meet different practical demands, such as comprehending different inputs (e.g., news~\cite{fabbri-etal-2019-multi} and dialogue~\cite{DBLP:conf/aaai/ZhongLX0022}) and generating different outputs (e.g., headlines~\cite{zhang-tetreault-2019-email} and paragraphs~\cite{perez-beltrachini-lapata-2021-models}).
Because annotating gold summaries for newly-proposed summarization tasks is costly~\cite{sen2008collective, zhang2022macsum}, few-shot summarization, the task of building a model for a specific summarization scenario using very limited ground-truth data~\cite{DBLP:conf/aaai/ChenS21a}, has gained increasing attention from the research community~\cite{ fabbri-etal-2021-improving, logan-iv-etal-2022-cutting, liu-etal-2022-psp, he2022z}.

\begin{figure}
    \centering
    \includegraphics[width=0.9\columnwidth]{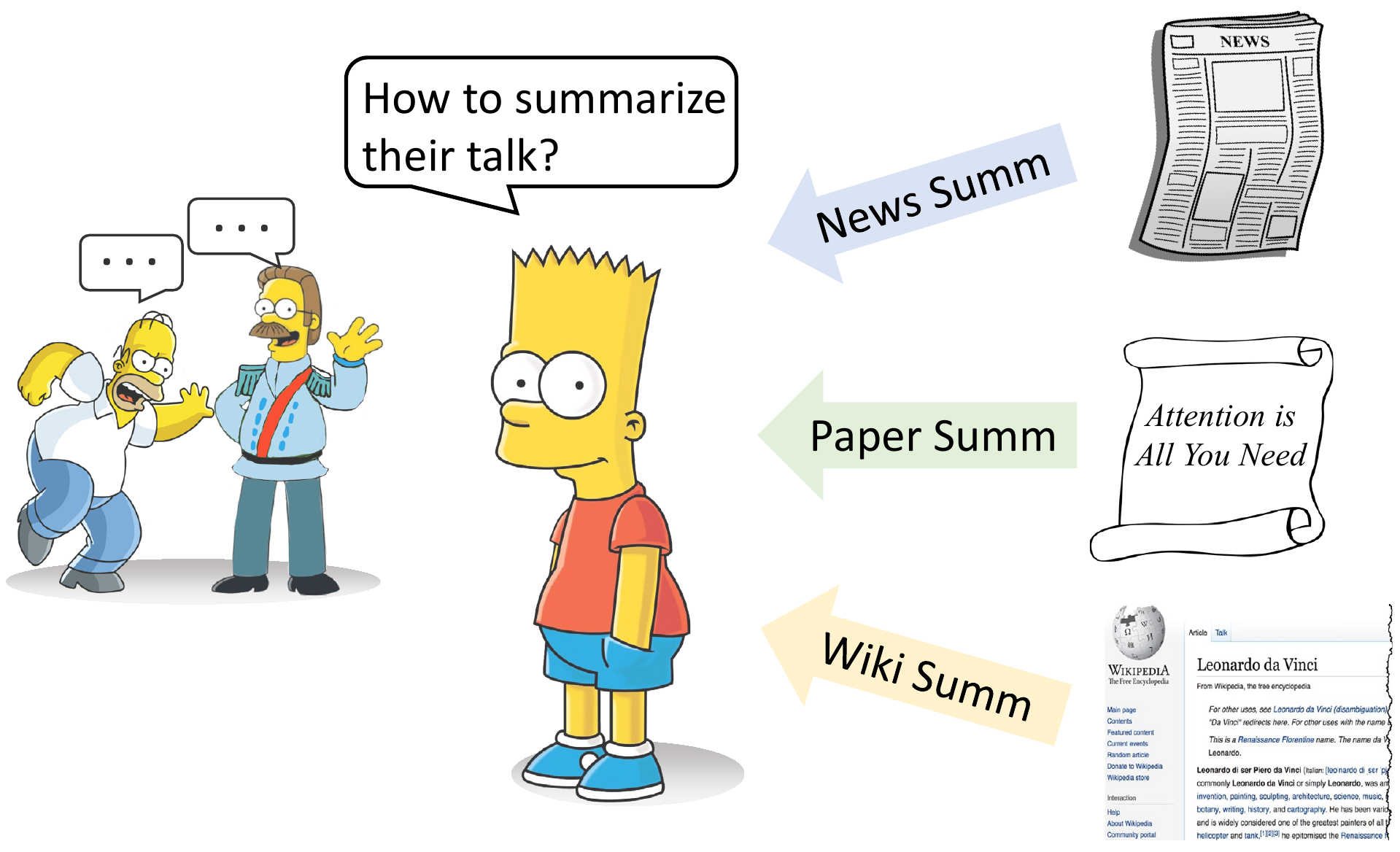}
    \caption{The few-shot summarization scenario in this paper. We are interested in how to re-use previous datasets (e.g., CNNDM) to improve the few-shot performance on unseen target tasks (e.g., \textsc{DialogSum}).}
    \label{fig:setting}
\end{figure}
Recently, prefix-tuning~\cite{li-liang-2021-prefix} has established strong baselines on many few-shot natural language generation tasks, including summarization.
The main idea is to extract knowledge from PLMs by prepending and tuning additional parameters (prefixes) before each layer of the PLM.
Work has been done to improve the performance by designing more sophisticated prefixes~\cite{ghazvininejad-etal-2022-discourse, liu-etal-2022-psp}.
Despite being effective, PLMs can have limited summarization knowledge due to the salient gap between pre-training objectives (e.g., language modeling) and summarization objectives~\cite{DBLP:conf/iclr/AribandiTSRZMZ022}.
In addition, existing summarization datasets can provide relevant knowledge to newly-proposed summarization tasks, and therefore benefit summarization tasks, especially under the few-shot scenario.
However, existing work tends to tune PLMs directly on a new task, without exploiting cross-task knowledge from summarization datasets,  which may limit the generalization and adaptation abilities of models~\cite{zhong2019closer, chen-yang-2021-structure, fang-etal-2022-spoken}.
We address these issues by proposing 
a unified few-shot summarization framework, \textbf{\textsc{UniSumm}}.
The idea is to combine multi-task pre-training~\cite{DBLP:conf/aaai/ChenS21a} on existing summarization datasets with few-shot prefix-tuning~\cite{li-liang-2021-prefix} on target tasks.
To this end, we first build a multi-task model based on a Transformer-based language model as the backbone and equip it with task-specific prefix vectors, and then  pre-train the multi-task model on diverse summarization datasets.
In this stage, we optimize the summarization model together with task-specific prefixes and also a \emph{universal prefix}, using an \emph{asymmetrical weight decay} strategy.
Using prefixes in the multi-task pre-training stage leads to two advantages: First, the mixture of shared summarization parameters and unique task-specific parameters helps to leverage natural benefits across datasets~\cite{ruder2017overview}. 
Second, the pre-trained prefixes can be tuned to serve as a knob for the second stage of prefix-tuning on unseen tasks.
When facing an unseen few-shot summarization  task, we freeze the multi-task learned backbone model and use the universal prefix as initialization for prefix-tuning.
A data obstacle for few-shot summarization research is the lack of a benchmark for fair comparison.
Previous studies either focus on one type of data, e.g., news text~\cite{liu-etal-2022-psp}, or train their systems on non-public few-shot samples. 
However, because few-shot models can be highly sensitive to training data, the selection of different few-shot samples in different papers 
can lead to ambiguous comparisons (a.k.a. \emph{Sample Selection Bias}~\cite{cortes2008sample}). 
To address these issues,
we assemble and release a new few-shot summarization benchmark, \textbf{\textsc{SummZoo}}, following two principles, namely \emph{diversity of tasks} and \emph{robustness of evaluation}.
\textsc{SummZoo} collects summarization data from 8 existing datasets, which are diverse in terms of domain (news, academic papers, meetings, etc.), format (single-document and multi-document), and length on both source and target sides.
For more robust evaluation, for each task, \textsc{SummZoo} provides 5 different (randomly sampled) few-shot training sets, and requires all systems to report their averaged results.
Finally, \textsc{SummZoo} includes 10-shot and 100-shot settings.

We compare \textsc{UniSumm} against several strong baselines, including a GPT-3.5 model (\texttt{text-davinci-002})~\cite{DBLP:conf/nips/BrownMRSKDNSSAA20,textdavinci002}, on \textsc{SummZoo} and conduct thorough analysis. 
Experimental results of automatic evaluation metrics show that \textsc{UniSumm} outperforms baselines across all sub-stasks and human evaluation shows that \textsc{UniSumm} achieves better performance than baselines of similar sizes and comparable performance compared with \texttt{text-davinci-002}.
Additionally, \textsc{UniSumm} is empirically found to be more stable and robust when facing different few-shot samples.
Analysis shows that combining multi-task pre-training and few-shot prefix-tuning is essential to the performance of \textsc{UniSumm} and other techniques, such as universal prefix and asymmetrical weight decay strategy, can all improve its generalization ability.
We release our code, model and benchmark at \url{https://github.com/microsoft/UniSumm}.

\begin{figure*}
    \centering
    \includegraphics[width=1\textwidth]{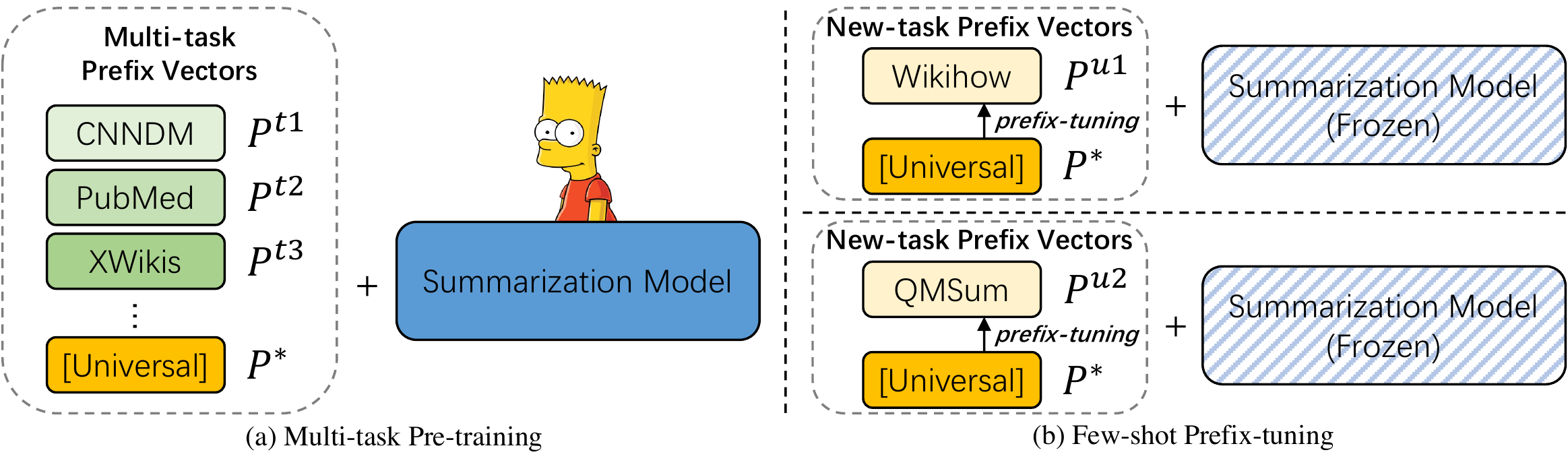}
    \caption{The two-phase framework of \textsc{UniSumm}.
    (a) The multi-task pre-training phase.
    The summarization model's parameters and prefixes are tuned together on multiple pre-training tasks, for example, \emph{CNNDM}, \emph{PubMed}, etc.
    (b) The few-shot tuning phase.
    For a new task, e.g., \emph{WikiHow}, we only tune prefix parameters while keeping the summarization model's parameters frozen. }
    \label{fig:framework}
\end{figure*}




\section{Related Work}\label{sec:related_work}
\paragraph{Few-shot Summarization}
A critical challenge for neural summarizers is that they are data-hungry and require large-scale annotated data.
To alleviate the data sparsity issue, \citet{fabbri-etal-2021-improving} extract characteristics of the target dataset and build pseudo summaries from the Wikipedia corpus.
Small plug-in networks \cite{brazinskas-etal-2020-shot} are injected into PLMs to predict the properties of the target dataset with only a small amount of labeled instances.
To close the gap between pre-training and fine-tuning, \citet{yu-etal-2021-adaptsum} propose a second stage of pre-training before fine-tuning with large-scale generative models.
Such challenges of summarization have also been explored in the cross-lingual setting \cite{bai-etal-2021-cross,chen-etal-2022-cross}.
Although transfer learning methods make use of external data, one still needs to carefully select source domains and tasks to avoid negative transfer~\cite{gururangan-etal-2020-dont,pilault2020conditionally}. 
Compared with them, \textsc{UniSumm} can be easily prefix-tuned to any target tasks without the effort of building large pseudo data or selecting relevant data.
To our knowledge, we are the first to combine prefix-tuning and multi-task learning for few-shot summarization, showing very positive results.

Existing few-shot summarization evaluation suffers from two data-related problems.
First, previous studies usually focus on only one type of summarization tasks in their experiments~\citep{brazinskas-etal-2020-shot,liu-etal-2022-psp}.
Thus, it is difficult to evaluate their generalization ability.
Second, the few-shot settings and selections of few-shot samples are miscellaneous, which makes evaluations from different research papers not comparable with each other~\cite{cortes2008sample}.
Therefore, in this work, we propose \textsc{SummZoo} for better benchmarking future research on few-shot summarization.
To our knowledge, \textsc{SummZoo} is the first public few-shot summarization benchmark that covers a set of diverse summarization tasks.

\paragraph{Prompt Learning for Text Generation}
The idea of prompt learning is first proposed in GPT-3~\citep{DBLP:conf/nips/BrownMRSKDNSSAA20}, where it aims to guide PLMs to do different tasks without further fine-tuning by prepending task-related examples to the input and has shown positive results on many text generation tasks, including summarization~\cite{goyal2022news}.
Prefix-tuning extends this idea from discrete tokens to continuous vectors~\cite{li-liang-2021-prefix}.
It adds continuous embeddings (prefixes) to each Transformer layer as external value and key vectors. 
During training, only prefixes are updated while the other parameters are unchanged.
\citet{logan-iv-etal-2022-cutting} and \citet{gu-etal-2022-ppt} propose to use pre-training to boost the low performance for few-shot learning. 
\citet{li-etal-2022-learning-transfer} combines the transfer learning and prompt learning for text generation.
Compared with them, we are interested in few-shot summarization and propose multi-task pre-training as an effective strategy to make use of data from related tasks to improve performance of diverse target tasks, which suits real-life scenarios.

\section{Method}\label{method}

Following~\citet{DBLP:conf/aaai/ChenS21a}, the task of \emph{{few-shot text summarization}} is defined as follows.
For an unseen target summarization task $u$, few-shot text summarization is to generate a summary $Y$, given an input text $X$, by learning from a limited number $k$ ($k \leq 100$ typically) of labeled training instances of $u$, with the help of general knowledge $K$.

The overall framework of \textsc{UniSumm} is shown in Figure~\ref{fig:framework}.
It consists of 2 phases: 1) Learning general knowledge by multi-task pre-training on existing summarization datasets (\S~\ref{sec:summarization_pre-training}) and; 2) Learning target task knowledge by prefix-tuning on each target few-shot summarization dataset (\S~\ref{sec:prefix-tuning}).

\subsection{Multi-Task Pre-Training with Prefix }\label{sec:summarization_pre-training}


As shown in Figure~\ref{fig:framework} (a),
in the first stage, 
we take a Transformer-based pre-trained language encoder-decoder model (for example, BART~\cite{lewis-etal-2020-bart}) $M = [M_{en}; M_{de}]$ as the summarization model, parameterized by $\theta$.
We further pre-train this  model on a set of popular summarization datasets (e.g., \emph{CNNDM},  \emph{PubMed} and \emph{XWikis}) to learn general summarization knowledge.
For each task $t$, we inject task-specific prefix vectors of encoder ($P^t_{en}$) and decoder ($P^t_{de}$), $P^t=[P^t_{en};P^t_{de}]$, into the model, parameterized by $\theta_{p^t}$.
Following \citep{li-liang-2021-prefix}, the prefix vectors are prepended to  each Transformer layer of $M$ as additional key and value vectors as: $[P^t_{en};M_{en};P^t_{de};M_{de}]$.



For all pre-training tasks, given input text $X$, the multi-task optimization objective is to minimize the negative log-likelihood of generating the target summary  $Y = \{y_1, y_2, ... y_{|Y|}\}$:
\begin{equation}
\begin{split}
L(\theta, \theta_{p^t}) =\sum^{|Y|}_{i}\log \mathbb{P}(&y_i|X, y_1,\cdots,y_{i-1}).
    \label{eq:prefix tuning}
\end{split}
\end{equation}

In the multi-task pre-training stage, we optimize ${\theta}$ and ${\theta_{p^t}}$ together.

\begin{table*}
\renewcommand{\arraystretch}{1.25}
    \centering
    \small
    \begin{tabular}{c|c|c|c|c|c}
    \thickhline
    \multicolumn{2}{c|}{\textbf{Type}}  &  \makebox[0.12\textwidth][c]{\textbf{Domain}}  & \makebox[0.12\textwidth][c]{ \textbf{Dataset}}    &\makebox[0.095\textwidth][c]{ \textbf{Testset Size}}  & \makebox[0.12\textwidth][c]{ \textbf{Avg. D/S Length}}\\
    \thickhline
     \multirow{6}{*}{Monologue}   & Multi-doc & \multirow{3}{*}{News}  &  MultiNews~\cite{fabbri-etal-2019-multi}   &  $5,622$   &  $2,103/264$\\
    \cline{2-2}
    \cline{4-6}
     &\makecell[c]{Extreme\\single-doc}  & & XSum~\cite{narayan-etal-2018-dont}    &  $11,334$   & $431/20$ \\
    \cline{2-6}
    \cline{2-6}
     & Single-doc & Scientific Paper  &  ArXiv~\cite{cohan-etal-2018-discourse}  &  $6,440$  &  $4,938/220$\\
    \cline{2-6}
     & Single-doc&  Instructions  &  WikiHow~\cite{koupaee2018wikihow}    &  $6,000$  &  $580/62$\\
    \cline{2-6}
    \cline{2-6}
     & Single-doc &  \makecell[c]{Online Forum}  &  Reddit-TIFU~\cite{kim-etal-2019-abstractive}  &  $4,208$  &  $433/23$\\
    \hline
    \multirow{4}{*}{Dialogue} & Single-doc & Online Chit-chat  &   SAMSum~\cite{gliwa-etal-2019-samsum}    &  $819$  &  $94/28$\\
    \cline{2-6} &Single-doc 
     &  Real-life &  \textsc{DialogSum}~\cite{chen-etal-2021-dialogsum}  &  $500$  &  $131/24$\\
    \cline{2-6} &\makecell[c]{Query-based\\single-doc}
     &  \makecell[c]{Meeting}   &  QMSum~\cite{zhong-etal-2021-qmsum}    &  $279$  &  $1,310/65$ \\

    \thickhline
    \end{tabular}
    \caption{Summary of sub-tasks in \textsc{SummZoo}.
    We report the sizes of test sets here. 
    ``Avg. D/S length'' stands for ``averaged document/summary token length''. 
    For QMSum, we concatenate the query and gold span as input.
    }
    \label{tab:benchmark}
\end{table*}

\subsection{Prefix-Tuning}\label{sec:prefix-tuning}
Through multi-task pre-training, we obtain the \textsc{UniSumm} model with diverse summarization knowledge. 
As shown in Figure~\ref{fig:framework} (b), for an unseen summarization task $u$ (for example, \emph{Wikihow} or \emph{MultiNews}),
given only $k$ training samples, we conduct prefix-tuning~\cite{li-liang-2021-prefix} on the \textsc{UniSumm} model.
A new-task prefix $P^u=[P^u_{en};P^u_{de}]$ is created, parameterized by $\theta_{p^u}$,  which can be either initialized randomly or from a prefix of pre-training tasks.
We then freeze the parameters $\theta$ of the shared summarization model and only tune $\theta_{p^u}$ using the objective defined in Equation~\ref{eq:prefix tuning}.
By doing this, we can maximize the learned summarization knowledge in \textsc{UniSumm} and also avoid over-fitting the model to very few samples.


\subsection{Universal Prefix}\label{sec:prefix-pretraining}
Empirically, given a target task, initializing  new-task prefix from the most related pre-training tasks can be helpful.
However, for a brand new task, selecting meta tasks can be a complicated process, which requires large efforts of feature engineering~\cite{DBLP:conf/aaai/ChenS21a}.
Therefore, during multi-task pre-training, we also pre-train a universal prefix, which can be used as a stable initialization for few-shot prefix-tuning.

In particular, during multi-task pre-training~(\S~\ref{sec:summarization_pre-training}), we initialize a universal encoder and decoder prefix vector  $P^*=[P^*_{en};P^*_{de}]$, parameterized by $\theta_{p^*}$.
For each training instance from task $t$, it has a $15\%$ probability to be coupled with this universal prefix vector instead of its task-specific prefix $P^t$.
The parameters $\theta_{p^*}$ are optimized together with $\theta$.
Then in prefix-tuning, we use this universal vector as initialization for the unseen task parameter $\theta_{p^u}$ (\S~\ref{sec:prefix-tuning}).

\subsection{Asymmetrical Weight Decay}\label{sec:weight_decay}
A potential problem in multi-task learning is the  negative transfer among different pre-training tasks. 
To alleviate this, inspired by previous work~\cite{evgeniou2004regularized,bengio2012practical, liu-etal-2019-single}, we set different weight decay regularizations on different parameters of \textsc{UniSumm}.
Specifically, we separate optimizers of the prefixes and the summarization model in pre-training.
We assign a lower weight decay value $d_p$=$0.01$ on the prefix optimizer, enabling prefixes to flexibly learn task-specific knowledge, and a higher weight decay value $d_l$=$0.05$ on the summarization model optimizer, enforcing it to learn a broader generalization across different tasks.

Formally, at training step $i$:
\begin{equation}
\begin{split}
    \theta^{i+1} & = (1-d_{l})\theta^i -\alpha^i\nabla f^i(\theta^i),\\
    \theta^{i+1}_p & = (1-d_{p})\theta^i_p-\alpha^i_p\nabla f^i_p(\theta^i_p),
\end{split}
\end{equation}

\noindent where $\alpha^i$ and $\alpha^i_p$ are the learning rates for summarization model and prefix, and $\nabla f^i(\theta^i)$ and $\nabla f^i_p(\theta^i_p)$ are the batch gradient for summarization model and prefix.

\section{The \textsc{SummZoo} Benchmark}\label{sec:the_summzoo_benchmark}


\textsc{SummZoo} is sourced from from existing summarization benchmark based on the principles of diversity and robustness, where we assemble each dataset into few-shot evaluation settings.

\paragraph{Diversity of Tasks}
As a major goal, we ensure that \textsc{SummZoo} can include a diversity of different summarization tasks, covering multiple domains, text styles and compression ratios.
Thus, we carefully select $8$ summarization tasks including  monologue/dialogue texts and single/multi-document summarization tasks. Their domains also span an assorted set such as news, scientific papers, instructions, online forums and meetings.


\paragraph{Robustness of Evaluation}
Our second goal is to ensure that experiments on \textsc{SummZoo} can be compared with each other in a robust manner. 
Also, we want to reduce the randomness from different selections of few-shot samples.
Therefore, for each task, we provide 5 sets of few-shot training samples, and we ask all models to  train on these 5 sets respectively and report their averaged results and standard deviations.
We also formulate two few-shot training settings with the number of shots $k$ set to $10$ or $100$, where the first can be considered as a more extreme low-resource scenario while the second is a more commonly tested setting.

Table~\ref{tab:benchmark} summarizes the statistics of sub-datasets in \textsc{SummZoo}. The detailed descriptions of each dataset can be found in Appendix~\ref{appendix:datasets_in_summzoo}.

\section{Experimental Setup}\label{sec:setup}

\subsection{Training Datasets}\label{sec:training_datasets}

For multi-task pre-training (\S~\ref{sec:summarization_pre-training}), we use a combination of seven summarization datasets: CNNDM~\cite{CNNDM}, BillSum~\cite{kornilova-eidelman-2019-billsum}, {PubMed}~\cite{cohan-etal-2018-discourse}, {GovReport}~\cite{huang-etal-2021-efficient}, {MediaSum}~\cite{zhu-etal-2021-mediasum}, {SummScreen}~\cite{chen-etal-2022-summscreen} and {XWikis}~\cite{perez-beltrachini-lapata-2021-models}.

To balance the training data size of different datasets, we perform down-sampling on over-sized datasets and up-sampling on low-resource datasets respectively. The detailed descriptions of each dataset and statistics of resulting data for pre-training are shown in Appendix~\ref{appendix:multi-task_pre-training_datasets} and Table~\ref{tab:training_data}.

\begin{table*}[ht]
\small
\setlength\tabcolsep{3pt}
\renewcommand{\arraystretch}{1.1}
\begin{tabular}{lr|ccc|ccc|ccc|ccc|ccc}
\thickhline

\multicolumn{2}{c|}{\multirow{2}{*}{\textbf{Task}}} & \multicolumn{3}{c|}{\textbf{PEGASUS}} & \multicolumn{3}{c|}{\textbf{BART-FT}} & \multicolumn{3}{c|}{\textbf{BART-PT}} & \multicolumn{3}{c|}{\textbf{MultiBART}} & \multicolumn{3}{c}{\textbf{\textsc{UniSumm}}} \\
\multicolumn{2}{c|}{} & \multicolumn{1}{c}{\textsc{R1}} & \multicolumn{1}{c}{R2} & \multicolumn{1}{c|}{RL} & \multicolumn{1}{c}{\textsc{R1}} & \multicolumn{1}{c}{R2} & \multicolumn{1}{c|}{RL} & \multicolumn{1}{c}{\textsc{R1}} & \multicolumn{1}{c}{R2} & \multicolumn{1}{c|}{RL} & \multicolumn{1}{c}{\textsc{R1}} & \multicolumn{1}{c}{R2} & \multicolumn{1}{c|}{RL} & \multicolumn{1}{c}{\textsc{R1}} & \multicolumn{1}{c}{R2} & \multicolumn{1}{c}{RL} \\ 
\thickhline
\multirow{2}{*}{MN} 
& \ \ 10 
& 39.12 & 11.15 & 19.44 
& 38.29 & 10.05 & 18.32 
& 38.27 & 11.38 & 19.28 
& 42.31 & 14.55 & 21.53 
& \textbf{45.13} & \textbf{15.19} & \textbf{21.63} \\
& 100 
& 42.36 & 12.78 & 20.56
& 42.65 & 13.27 & 20.69 
& 43.86 & 13.97 & 20.79
& 45.71 & 15.78 & 22.21 
& \textbf{45.91} & \textbf{15.86} & \textbf{22.24} 
\\ \hline
\multirow{2}{*}{XSum}
& \ \ 10 
& 20.55 & \ \ 3.98 & 14.80 
& 24.89 & \ \ 6.42 & 19.18
& 14.29 & \ \ 2.77 & 11.52 
& 20.76 & \ \ 5.76 & 17.01 
& \textbf{26.10} & \ \ \textbf{7.20} & \textbf{19.92} \\
& 100 
& \textbf{37.30} & \textbf{13.69} & \textbf{29.08} 
& 27.45 &\ \  7.21 & 21.74 
& 29.70 & \ \ 9.87 & 23.70 
& 31.48 & 10.88 & 25.00 
& \textbf{33.33} & \textbf{11.36} & \textbf{25.85}
\\ \hline
\multirow{2}{*}{ArXiv}
&\ \ 10 
& 34.81 &\ \ 8.46 & 29.12 
& 28.40 &\ \ 4.98 & 25.15
& 29.85 &\ \ 8.08 & 26.76 
& 41.45 & 14.68 & 37.01 
& \textbf{43.33} & \textbf{15.38} & \textbf{38.69} \\
& 100 
& 38.08 & 10.14 & 31.06 
& 36.69 & 10.07 & 32.67
& 38.03 & 11.46 & 34.20 
& 43.56 & 15.97 & 39.01 
& \textbf{44.33} & \textbf{16.42} & \textbf{39.71} 
\\ \hline
\multirow{2}{*}{WH}
&  \ \ 10 
& 27.74 & \ \  7.80 & 19.61 
& 17.09 & \ \ 2.37 & 12.01 
& 25.31 & \ \ 7.45 & 19.02 
& 27.64 & \ \ 7.99 & 19.91 
& \textbf{30.87} &\ \ \textbf{9.35} & \textbf{21.72} \\
& 100 
& 33.21 & 10.86 & 24.41 
& 26.46 &\ \ 6.91 & 18.83 
& 32.35 & 10.42 & 23.23 
& 34.10 & 11.31 & 25.03 
& \textbf{34.90} & \textbf{11.73} & \textbf{25.70} 
\\ \hline
\multirow{2}{*}{Reddit}
& \ \ 10 
& 18.90 &\ \ 3.89 & 14.27 
& 13.80 &\ \ 1.20 & 10.48 
& 19.01 &\ \ 4.07 & 14.46 
& 21.44 &\ \ 5.17 & 16.22 
& \textbf{22.88} &\ \  \textbf{5.60} & \textbf{17.02} \\
& 100 
& 23.40 & \ \ 5.71 & 17.99 
& 17.91 &  \ \ 2.58 & 13.33 
& 23.10 & \ \ 5.41 & 17.42 
& 24.06 &\ \ 5.89 & 17.97 
& \textbf{24.54} &\ \ \textbf{6.17} & \textbf{18.30} 
\\ \hline
\multirow{2}{*}{\textsc{DS}}
& 10 
& 36.44 & 10.89 & 28.49 
& 28.62 & \ \ 5.97 & 22.83 
& 33.46 & 10.08 & 27.90 
& 37.05 & 12.61 & 30.24 
& \textbf{38.76} & \textbf{13.38} & \textbf{31.07} \\
& 100
& 41.02 & 14.53 & 32.29 
& 38.77 & 12.91 & 31.40 
& 41.20 & 13.97 & 32.76 
& 42.16 & \textbf{15.71} & \textbf{33.79}
& \textbf{42.43} & 15.64 & 33.74
\\ \hline
\multirow{2}{*}{SS}
& 10 
& 38.58 & 13.79 & 30.37 
& 18.07 & \ \ 4.23 & 14.70 
& 35.53 & 12.96 & 28.26 
& 39.69 & 16.28 & 32.11 
& \textbf{43.89} & \textbf{18.53} & \textbf{34.76} \\
& 100 
& 44.60 & 18.40 & 35.16
& 37.36 & 14.14 & 30.02
& 43.39 & 17.82 & 34.42
& 45.47 & 19.68 & 36.60 
& \textbf{46.93} & \textbf{20.65} & \textbf{37.28} 
\\ \hline
\multirow{2}{*}{QM}
& 10 
& 31.77 & \ \ 9.70 & 21.48
& 23.64 & \ \ 3.56 & 14.88 
& 27.58 & \ \ 8.39 & 19.41 
& 33.71 & 10.59 & 22.27 
& \textbf{36.00} & \textbf{12.12} & \textbf{23.56} \\
& 100 
& 35.54 & 11.68 & 23.74 
& 33.96 & 10.30 & 22.10 
& 35.07 & 11.66 & 23.10 
& 37.67 & 13.38 & 24.68 
& \textbf{38.38} & \textbf{13.89} & \textbf{25.36} 
\\ \hline
\multirow{2}{*}{Average}
& 10 
& 30.99 &\ \ 8.71 & 22.20 
& 24.10 &\ \ 4.85 & 17.19 
& 27.91 & \ \ 8.15 & 20.83 
& 33.01 & 10.95 & 24.54
& \textbf{35.87} & \textbf{12.09} & \textbf{26.05} \\
& 100 
& 36.94 & 12.22 & 26.79 
& 32.66 &\ \ 9.67 & 23.85
& 35.84 & 11.82 & 26.20
& 38.03 & 13.58 & 28.04 
& \textbf{38.84} & \textbf{13.97} & \textbf{28.52} 
\\ \thickhline
\end{tabular}
    \caption{Main results of PEGASUS, BART-FT, BART-PT, MultiBART and \textsc{UniSumm} on the \textsc{SummZoo} benchmark. 
    MN, WH, DS, SS and QM are abbreviations of MultiNews, WikiHow, \textsc{DialogSum}, SAMSum and QMSum. Best results on each sub-dataset are in bold.    
    All models are trained on the same 5 sets of few-shot samples and we report their averaged \textsc{Rouge} scores.
    The bottom block presents the averaged results of all 8 sub-tasks in \textsc{SummZoo}.
    }
    \label{tab:main_results}
\end{table*}


\subsection{Baseline Models}

\paragraph{PEGASUS}~\cite{DBLP:conf/icml/ZhangZSL20} is a large pre-trained encoder-decoder model, which is particularly designed for text summarization.
The model is trained using the gap sentence generation task.
We use $\textsc{PEGASUS}_{\textsc{large}}$ (C4+HugeNews)\footnote{\url{https://huggingface.co/google/pegasus-large}} for comparison, which improves upon the results reported in the original paper.

\paragraph{BART}~\cite{lewis-etal-2020-bart} is a pre-trained encoder-decoder language model using self-denoising tasks.
We compare with the BART-large model\footnote{\url{https://huggingface.co/facebook/bart-large}} with two tuning strategies on few-shot summarization tasks, namely standard fine-tuning (\textbf{BART-FT}) and prefix-tuning (\textbf{BART-PT}). In BART-PT, the prefix vector is added in the same way as in \textsc{UniSumm}.

\paragraph{MultiBART} is a variant of BART-large. Similar to \textsc{UniSumm}, it is first multi-task pre-trained on the \textbf{\emph{same data}} (\S~\ref{sec:training_datasets}) but \emph{without} prefixes.
And it can also be fine-tuned or prefix-tuned to fit few-shot summarization tasks.
We only show the results of prefix-tuned MultiBART because we find fine-tuning the entire MultiBART model always leads to worse performance in the few-shot setting.
This strong baseline can be considered as an indicator to verify the effectiveness of using prefixes in both multi-task pre-training and few-shot tuning.

\paragraph{\texttt{Text-davinci-002}}~\cite{DBLP:conf/nips/BrownMRSKDNSSAA20, textdavinci002} is a large language model ($175$B) from the GPT-3.5 family,\footnote{\url{https://openai.com/}} using instruction tuning, and has shown great zero-/few-shot performance on many NLP tasks, including summarization.
Specifically, recent work finds that GPT-3.5 models can show much better performance with the technique of in-context learning (ICL)
~\cite{DBLP:conf/nips/BrownMRSKDNSSAA20,liu2022few}.
We use \texttt{text-davinci-002} with ICL for experiments, and only show the performance of $1$-shot ICL because of its input length limitation.\footnote{For MultiNews and ArXiv, due to the length limitation of GPT-3.5 API, we only include the summary part in their ICL examples.}

All baseline models and \textsc{UniSumm} are evaluated on \textsc{SummZoo} (Appendix~\ref{appendix:Implementation_details} shows the implementation details).
We conduct both automatic and human evaluation.
As described, \textsc{SummZoo} requires models to report averaged results and their standard deviations over $5$ sets of different few-shot samples (except for \texttt{text-davinci-002}).
We use \textsc{Rouge}~\cite{lin-2004-rouge} for automatic evaluation\footnote{{We use the \href{https://github.com/pltrdy/files2rouge}{files2rouge} for evaluation.}},
which evaluates the $n$-gram overlap in the model-generated summary against the reference summary.
We report the $F$-1 scores of \textsc{Rouge-1} (\textsc{R1}), \textsc{Rouge-2} (R2) and \textsc{Rouge-L} (RL).

\begin{table}[t]
    \setlength\tabcolsep{3.5pt}
    \renewcommand{\arraystretch}{1.1}
    \centering
    \small
    \begin{tabular}{l|c|c|c}
    \thickhline
        {\textbf{Task}} &   \makebox[0.08\textwidth][c]{\textbf{GPT-3.5}}
 & \makebox[0.08\textwidth][c]{\textbf{10-\textsc{Uni}}} & \makebox[0.08\textwidth][c]{\textbf{100-\textsc{Uni}}}\\
        \thickhline
        MultiNews
        & 11.01 & 15.19 & 15.86\\
        \hline
        Xsum & \ \ 8.87 &  \ \ 7.20 & 11.36 \\
        \hline
        Arxiv & 10.83 & 15.38 & 16.42 \\
        \hline
        WikiHow &\ \ 8.56 &\ \ 9.35 & 11.73\\
        \hline
        Reddit & \ \ 6.03 &\ \ 5.60 &\ \ 6.17 \\
        \hline
        \textsc{DialogSum} &13.08 & 13.38 & 15.64 \\
        \hline
        SAMSum &  17.65 & 18.53 & 20.65\\
        \hline
        QMSum & 11.62 & 12.12 & 13.89 \\\hline
        Average &  10.96 &12.09 & 13.97  \\
        \thickhline

    \end{tabular}
    \caption{\textsc{R2} scores of $1$-shot \texttt{text-davinci-002} (GPT-3.5) using ICL compared with 10-shot \textsc{UniSumm} and 100-shot \textsc{UniSumm}. }
    \label{tab:gpt3}
\end{table}

\section{Automatic Evaluation}\label{sec:results}

\subsection{Main Results}\label{sec:main_results}
The main results are shown in Table~\ref{tab:main_results} and \ref{tab:gpt3}.
First, compared with PEGASUS, \textsc{UniSumm} outperforms it across all tasks except 100-shot XSum, and shows the best averaged scores in both 10-shot and 100-shot settings.
We also find that 10-shot \textsc{UniSumm} can outperform 100-shot PEGASUS on MultiNews, Arxiv and QMSum by a large margin, suggesting that \textsc{UniSumm} can benefit from diverse training data and effectively adapt indirect knowledge to unseen tasks.
It is notable that although the foundation BART model is inferior to PEGASUS, the BART-based \textsc{UniSumm} can still outperform PEGASUS with the learned summarization knowledge.
Overall, \textsc{UniSumm} surpasses both BART-FT and BART-PT by a large margin on all tasks in all settings, which suggests the equipment of multi-task learning can substantially improve model performance on few-shot summarization tasks, in particular in the 10-shot setting.

\textsc{UniSumm} also outperforms MultiBART by a large margin, especially in the 10-shot setting (Avg. $2.86$ \textsc{R1} improvements).
Considering that MultiBART is multi-task pre-trained on the exact same data as \textsc{UniSumm} does, the main difference from \textsc{UniSumm} is whether to use prefixes in both multi-task pre-training and few-shot tuning.
The result verifies the effectiveness of \textsc{UniSumm} framework, in particular the prefix addition in the multi-task pre-training phrase (\S~\ref{sec:summarization_pre-training}).




The comparison between \texttt{text-davinci-002} and \textsc{UniSumm} is shown in Table~\ref{tab:gpt3}.
Generally, 100-shot \textsc{UniSumm} achieves higher \textsc{Rouge} scores than 1-shot \texttt{text-davinci-002} on all tasks and overall performance and 10-shot \textsc{UniSumm} shows better performance compared with 1-shot \texttt{text-davinci-002} except for XSum and Reddit.
Such improvements can be attributed to the fact that \textsc{UniSumm} is few-shot trained on more samples.
It is also worth noting that \textsc{UniSumm} is based on BART-large (400M), while GPT-3.5 is orders of magnitude larger (175B).
Also, we note that 10-shot \textsc{UniSumm} can achieve higher \textsc{Rouge} scores on some tasks such as MultiNews and Arxiv compared with \texttt{text-davinci-002}.
Besides \textsc{UniSumm} is multi-task trained on relevant data, one possible reason is that \texttt{text-davinci-002} is only presented with 1-shot summary as ICL context, due to the length limitation.
However, given the previous finding~\cite{goyal2022news} that GPT-3.5 generated summaries can be favored by human evaluators with even lower \textsc{Rouge} scores, we also conduct human evaluation in \S~\ref{sec:human_evaluation}.

\begin{table}[]
    \renewcommand{\arraystretch}{1.1}
\small
\centering

\begin{tabular}{lr|cccc}
\thickhline
\multicolumn{2}{c|}{\textbf{Task}}             & \textbf{PEG}           & \textbf{B-PT}          & \textbf{Mul}           & \textbf{\textsc{Uni}}           \\ \thickhline
\multirow{2}{*}{MultiNews} & 10  & 0.37          & 1.04          & 0.68          & \textbf{0.33} \\
                           & 100 & 0.20           & \textbf{0.11} & 0.26          & 0.19          \\ \hline
\multirow{2}{*}{XSum}      & 10  & 1.45          & 1.60           & 1.65          & \textbf{1.21} \\
                           & 100 & 0.37          & 0.27          & \textbf{0.11} & 0.27          \\  \hline
\multirow{2}{*}{Arxiv}     & 10  & 0.57          & 1.08          & \textbf{0.32} & 0.93          \\
                           & 100 & 0.55          & 0.83          & 0.64          & \textbf{0.54} \\ \hline
\multirow{2}{*}{WikiHow}   & 10  & 0.79          & 0.66          & 0.66          & \textbf{0.40}  \\
                           & 100 & 0.46          & 0.25          & 0.38          & \textbf{0.21} \\ \hline
\multirow{2}{*}{Reddit}    & 10  & \textbf{0.83} & 1.61          & 1.20           & 1.16          \\
                           & 100 & 0.71          & 0.72          & 0.68          & \textbf{0.52} \\ \hline
\multirow{2}{*}{\textsc{DialogSum}} & 10  & 1.18          & \textbf{0.96} & 1.46          & 0.99          \\
                           & 100 & \textbf{0.83} & 0.90           & 1.01          & 0.91          \\ \hline
\multirow{2}{*}{SAMSum}    & 10  & 1.61          & 1.58          & 1.91          & \textbf{1.07} \\
                           & 100 & 0.47          & \textbf{0.29}          & 0.40  & 0.47          \\ \hline
\multirow{2}{*}{QMSum}     & 10  & 0.84          & 0.75          & 0.71          & \textbf{0.45} \\
                           & 100 & 0.72          & 0.55          & 0.34          & \textbf{0.30}  \\ \hline
\multirow{2}{*}{Average}   & 10  & 0.96          & 1.16          & 1.07          & \textbf{0.82} \\
                           & 100 & 0.54          & 0.49          & 0.48          & \textbf{0.43} \\ \thickhline
\end{tabular}
    \caption{Comparison of model robustness towards different few-shot samples. 
    We report the \emph{standard deviations} of \textsc{R1} scores on $5$ sets of few-shot samples provided in \textsc{SummZoo}. 
    Lower standard deviation indicates the model is more robust towards different few-shot samples.
    The bottom block shows the averaged results of all 8 sub-tasks.
    Models are PEGASUS (PEG), BART-PT (B-PT), MultiBART (Mul) and \textsc{UniSumm} (\textsc{Uni}).
    } \label{tab:robustness}
\end{table}

\subsection{Model Robustness}
The sample selection bias~\cite{cortes2008sample} has been a major problem for few-shot tasks, where model performance is strongly correlated with the selection of few-shot samples. 
And a sound system should be robust and stable when taking different few-shot samples.
To demonstrate the robustness and stability of different few-shot summarization models, we report their standard deviations of \textsc{Rouge-1} scores on $5$ different sets of few-shot samples provided in \textsc{SummZoo} in Table~\ref{tab:robustness}.

Overall, the standard deviations of \textsc{UniSumm} are lower than all other baselines on most tasks in both settings, suggesting that \textsc{UniSumm} is most stable and robust when facing different few-shot samples.
Also, MultiBART outperforms BART-PT and shows better averaged results than PEGASUS in the 100-shot, showing that reusing related summarization datasets is valuable.
However, it can still be unstable in the 10-shot setting.
In contrast, \textsc{UniSumm} shows the least averaged standard deviations across all tasks in both settings.
This suggests that the two-phase training with prefixes in the \textsc{UniSumm} framework is essential for enhancing the model robustness.

We present the full table, including standard deviations of \textsc{R2} and \textsc{RL} scores, in Appendix~\ref{appendix:model_robustness}.
Overall, we find that \textsc{UniSumm} is most robust and stable towards different training samples.

\begin{table}[t]
    \setlength\tabcolsep{2.5pt}
    \renewcommand{\arraystretch}{1.1}
    \centering
    \small
    \begin{tabular}{lr|ccccc}
    \thickhline
\multicolumn{2}{c|}{\textbf{Task}} &  \makebox[0.065\textwidth][c]{\textbf{Gold}} & \makebox[0.065\textwidth][c]{\textbf{GPT-3.5} }&  \makebox[0.065\textwidth][c]{\textbf{PEG}} &  \makebox[0.065\textwidth][c]{\textbf{B-PT}} &  \makebox[0.065\textwidth][c]{\textbf{\textsc{Uni}}}\\
\thickhline
\multirow{4}{*}{QM} & \emph{Flu.}& 4.80 & 4.93 & 4.46 & 4.40 & 4.90  \\
& \emph{Coh.} & 4.93 & 4.80 & 4.10 & 3.87 & 4.50 \\
& \emph{Con.} & 5.00 & 4.03 & 3.33 & 3.13 & 3.80 \\
& \emph{Rel.} & 4.90 & 4.17 & 3.27 & 2.80 & 3.97 \\
\hline
\multirow{4}{*}{WH} & \emph{Flu.}& 4.72 & 4.90 & 4.43 & 4.30 & 4.68 \\
& \emph{Coh.} & 4.57 & 4.83 & 4.17 & 4.00 & 4.43 \\
& \emph{Con.} & 4.87 & 4.63 & 4.17 & 3.93 & 4.67 \\
& \emph{Rel.} & 4.88 & 4.58 & 4.33& 4.17 & 4.67  \\
\hline
\multirow{4}{*}{MN} & \emph{Flu.} & 4.70 & 4.97 & 4.23 & 4.17 & 4.63 \\
& \emph{Coh.} & 4.70 & 4.73 & 3.95 & 3.80 & 4.17\\
& \emph{Con.} & 4.93 & 3.07 & 3.53 & 3.27 & 4.07  \\
& \emph{Rel.} & 4.77 & 2.73 & 3.72 & 3.63 & 4.30\\
\thickhline
\end{tabular}
\caption{Human evaluation for gold summaries, 1-shot \texttt{text-davinci-002} (GPT-3.5), 100-shot PEGASUS (PEG), BART-PT (B-PT) and \textsc{UniSumm} (\textsc{Uni}) on QMSum, WikiHow and MultiNews.
\emph{Flu.}, \emph{Coh.}, \emph{Con.} and \emph{Rel.} stand for \emph{Fluency}, \emph{Coherence}, \emph{Consistency} and \emph{Relevance}, respectively.}
\label{tab:human_eval}
\end{table}

\begin{table*}[]
\small
    \setlength\tabcolsep{2.5pt}
    \renewcommand{\arraystretch}{1.1} 
    \centering
\begin{tabular}{l|ll|ll|ll|ll|ll|ll|ll|ll|ll}
\thickhline
\multirow{2}{*}{\textbf{Model}} & \multicolumn{2}{c|}{\textbf{MN}}                           & \multicolumn{2}{c|}{\textbf{XSum}}                         & \multicolumn{2}{c|}{\textbf{Arxiv}}                        & \multicolumn{2}{c|}{\textbf{WH}}                           & \multicolumn{2}{c|}{\textbf{Reddit}}                       & \multicolumn{2}{c|}{\textbf{DS}}                           & \multicolumn{2}{c|}{\textbf{SS}}                           & \multicolumn{2}{c|}{\textbf{QM}}                        & \multicolumn{2}{c}{\textbf{Avg.}}                          \\ 
                      & \multicolumn{1}{c}{10} & \multicolumn{1}{c|}{100} & \multicolumn{1}{c}{10} & \multicolumn{1}{c|}{100} & \multicolumn{1}{c}{10} & \multicolumn{1}{c|}{100} & \multicolumn{1}{c}{10} & \multicolumn{1}{c|}{100} & \multicolumn{1}{c}{10} & \multicolumn{1}{c|}{100} & \multicolumn{1}{c}{10} & \multicolumn{1}{c|}{100} & \multicolumn{1}{c}{10} & \multicolumn{1}{c|}{100} & \multicolumn{1}{c}{10} & \multicolumn{1}{c|}{100} & \multicolumn{1}{c}{10} & \multicolumn{1}{c}{100} \\ \thickhline
3-Task & \textbf{15.3} &15.8 & \ \ 4.8 & 10.9 & 15.0 & 15.7 &  \ \ 9.2 & \textbf{11.9 }& \ \ \textbf{5.7} &\ \ 6.1 &  12.6 & 15.6 & 17.1 & 19.8 & 11.5 & 13.5 & 11.4 & 13.6          \\
7-Task &  15.2 & \textbf{15.9} & \ \ \textbf{7.2 }& \textbf{11.4} & \textbf{15.4} & \textbf{16.4} & \ \ \textbf{9.4} & 11.7 &\ \ 5.6 &\ \ \textbf{6.2} & \textbf{13.4} & \textbf{15.7} & \textbf{18.5} & \textbf{20.7}  &\textbf{12.1} & \textbf{13.9} & \textbf{12.1 }& \textbf{14.0}  \\ \thickhline               
\end{tabular}
\caption{\textsc{Rouge-2} results of \textsc{UniSumm} models which are multi-task pre-trained on different scale of pre-training tasks. We show the best results in \textbf{bold}.}
\label{tab:number_results}
\end{table*}

\begin{table*}[h]
\small
    \setlength\tabcolsep{2.5pt}
    \renewcommand{\arraystretch}{1.1} 
    \centering
\begin{tabular}{l|ll|ll|ll|ll|ll|ll|ll|ll|ll}
\thickhline
\multirow{2}{*}{\textbf{Prefix}} & \multicolumn{2}{c|}{\textbf{MN}}                           & \multicolumn{2}{c|}{\textbf{XSum}}                         & \multicolumn{2}{c|}{\textbf{Arxiv}}                        & \multicolumn{2}{c|}{\textbf{WH}}                           & \multicolumn{2}{c|}{\textbf{Reddit}}                       & \multicolumn{2}{c|}{\textbf{DS}}                           & \multicolumn{2}{c|}{\textbf{SS}}                           & \multicolumn{2}{c|}{\textbf{QM}}                        & \multicolumn{2}{c}{\textbf{Avg.}}                          \\ 
                      & \multicolumn{1}{c}{10} & \multicolumn{1}{c|}{100} & \multicolumn{1}{c}{10} & \multicolumn{1}{c|}{100} & \multicolumn{1}{c}{10} & \multicolumn{1}{c|}{100} & \multicolumn{1}{c}{10} & \multicolumn{1}{c|}{100} & \multicolumn{1}{c}{10} & \multicolumn{1}{c|}{100} & \multicolumn{1}{c}{10} & \multicolumn{1}{c|}{100} & \multicolumn{1}{c}{10} & \multicolumn{1}{c|}{100} & \multicolumn{1}{c}{10} & \multicolumn{1}{c|}{100} & \multicolumn{1}{c}{10} & \multicolumn{1}{c}{100} \\ \thickhline
Random & \textbf{15.6} & \textbf{16.0} & \ \ 4.4 & 11.1 & \textbf{16.2} & 16.3 & \ \ \textbf{9.4} & 11.6 & \ \ \textbf{6.0} & \ \ 6.1 & 13.3 & \textbf{15.7} & 18.1 & \textbf{21.0} & 11.9 & 13.7 & 11.9 & 13.9       \\
CNNDM & 15.1 & 15.8 & \ \ 6.3 & 11.1 & 14.8 & 15.8 & \ \ 9.4 & 11.7 & \ \ 5.6 &\ \ 6.1 & 13.1 & 15.5 & \textbf{18.7} & 20.7 & 11.9 & 13.7 & 11.9 & 13.8 \\
Universal & 15.2 & 15.9 & \ \ \textbf{7.2 }& \textbf{11.4} & 15.4 & \textbf{16.4} & \ \ 9.4 &  \textbf{11.7} & \ \ 5.6 &  \ \ \textbf{6.2 }&  \textbf{13.4} & 15.6 & 18.5 & 20.7 & \textbf{12.1} & \textbf{13.9} & \textbf{12.1} & \textbf{14.0}
\\ \thickhline               
\end{tabular}
    \caption{\textsc{Rouge-2} results of \textsc{UniSumm} using different prefix initialization strategies. We show the best results in \textbf{bold}.}
    \label{tab:prefix_embedding}
\end{table*}

\section{Human Evaluation}\label{sec:human_evaluation}
To better understand the outputs of different few-shot summarization systems,
following~\citet{kryscinski-etal-2019-neural, kryscinski-etal-2020-evaluating}, we conduct a human evaluation from four dimensions: \emph{Fluency}, \emph{Consistency}, \emph{Coherence} and \emph{Relevance}.
We select 30 samples from QMSum, WikiHow and MultiNews, respectively, covering both monologue and dialogue texts.
Then, for each sample, we ask a judge with experience in human evaluation for summarization tasks, to give scores from 1 to 5 (higher score indicates better quality) along each evaluation dimension. Candidate outputs are from gold summaries, 1-shot \texttt{text-davinci-002}, 100-shot PEGASUS, BART-PT and \textsc{UniSumm}  respectively.
In total, we have 450 summaries to evaluate and the results are reported in Table~\ref{tab:human_eval}.
Appendix~\ref{appendix:human_evaluation} gives detailed description of evaluation dimensions.

In human evaluation, \textsc{UniSumm} outperforms PEGASUS and BART-PT on all datasets regarding all dimensions, achieving a higher fluency score than gold summaries on QMSum and a comparable score on MultiNews and WikiHow, suggesting that \textsc{UniSumm} can generate very fluent sentences which can be comparable with human annotated summaries. 
A challenge of QMSum is that models are asked to generate summaries focusing on the input queries.
Thus, \emph{Relevance} is a very important metric for this task.
However, \emph{Relevance} sees very low score for PEGASUS ($3.27$) and BART-PT ($2.80$), suggesting they are weak in extracting relevant information based on user queries.
In contrast, \textsc{UniSumm} achieves a higher score ($3.97$).
\texttt{Text-davinci-002} also performs very well on this task, even outperforming the gold summaries on \emph{Fluency}, but \textsc{UniSumm} still achieves comparable results with limited training samples and much lower cost.

On MultiNews, since \texttt{text-davinci-002} is only input with 1-shot summary as ICL example due to length limitation, although it can generate very fluent ($4.97$) and coherent ($4.73$) summaries, it is less preferred by human annotators w.r.t. \emph{Consistency} and \emph{Relevance}.
\textsc{UniSumm} still outperforms other systems and only loses to gold summaries on this two metrics.
Similar results are also observed on WikiHow, where \texttt{text-davinci-002} tends to generate very long summaries, which can contain some hallucination and less important content, and \textsc{UniSumm} shows comparable performance on \emph{Consistency} and \emph{Relevance}.
We show case studies and their analysis, including an error case where \textsc{UniSumm} fails, in Appendix~\ref{appendix:case_study}. 



\section{Analysis}\label{sec:analysis}

\subsection{Task Scale in  Multi-task Training}\label{sec:task_scale}
One common concern about multi-task training is that: when multiple tasks are combined, will newly added tasks hurt or help the performance?
To verify this, we add one variant of \textsc{UniSumm} for comparison, whose phase-1 is multi-task pre-trained on $3$ tasks instead of all $7$ tasks in Table~\ref{tab:training_data}.
For the 3 tasks, we use the combination of CNNDM, PubMed and MediaSum, which are typical datasets for news summarization (MultiNews and Xsum), academic paper summarization (ArXiv) and dialogue summarization (\textsc{DialogSum}, SAMSum and QMSum).

Results in Table~\ref{tab:number_results} show that when extending the multi-task pre-training datasets from 3 to 7, \textsc{UniSumm} achieves better results on multiple datasets. 
For example, taking ArXiv as the target task, 7-Task \textsc{UniSumm} outperforms 3-Task \textsc{UniSumm} in both 10 and 100-shot settings.
It suggests that 7-Task \textsc{UniSumm} can benefit from  GovReport, XWikis, SummScreen and BillSum for scientific text summarization.
On average, the \textsc{R2} score improves by $0.4$ for the 10-shot setting and $0.7$ for the 100-shot setting.
This shows that negative transfer is minor in \textsc{UniSumm} and suggests that by training \textsc{UniSumm} on even more datasets, its generalization can potentially be improved by learning more indirect summarization knowledge.

\subsection{Different Prefix Initializations}\label{sec:general_prefix}
\textsc{UniSumm} is equipped with a universal prefix that was randomly (15\%) picked by all tasks during multi-task pre-training (\S~\ref{sec:prefix-pretraining}). 
In Table~\ref{tab:prefix_embedding}, we show the ablation study of using different prefix initialization strategies in few-shot prefix-tuning.
Due to space limitation, we show \textsc{R-2} scores here.
We compare three strategies: initialized the prefix randomly, using \emph{CNNDM} prefix or using universal prefix.
The \emph{CNNDM} prefix is selected to be compared here because it is considered as a general summarization task and has been proved helpful to many tasks, e.g., SAMSum~\cite{gliwa-etal-2019-samsum}.

We see that using universal prefix yields the best results on most tasks.
Also, the universal prefix is particularly useful for the 10-shot setting, bringing a $0.23$ improvement for \textsc{R2} score.
In addition, we find that using task-specific prefix (\emph{CNNDM}) shows the worst performance on some tasks, such as QMSum and ArXiv, and has the lowest average score.
This can be explained by that the task-specific prefix (\emph{CNNDM}) stores abundant task specific knowledge, which however can be harmful to unseen target tasks, especially when the target task is very different from the pre-training task.

We show more analysis in Appendix~\ref{appendix:influence_of_wedight_decay}.

\section{Conclusion}\label{sec:conclusion}
We introduced \textsc{UniSumm}, a novel few-shot summarization system that can be easily prefix-tuned to excel at and generalize on a diversity of summarization tasks.
We propose to combine multi-task learning and prefix-tuning by jointly training the prefixes and the summarizer on multiple existing summarization datasets. 
By only tuning the prefix parameters, \textsc{UniSumm} shows superior performance over strong baseline systems, yielding fluent and faithful summaries across tasks. 
In addition, we assembled and released a new benchmark, \textsc{SummZoo}, for fairly and effectively evaluating few-shot summarization models. 
It covers an assorted set of summarization tasks and provides multiple few-shot sets for a more robust and fairer comparison.



\section*{Limitations}
The limitation of \textsc{UniSumm} can be stated from three perspectives.
First, the multi-task pre-training of \textsc{UniSumm} can be time and cost consuming, which requires large GPU resources.
Second, the current framework uses prefixes of a fixed length for both multi-task training and few-shot prefix-tuning.
However, different summarization task may prefer different size of prefixes.
Third, in this work, we focus on summarization tasks in English.
The performance of \textsc{UniSumm} for languages that have a different morphology or syntactic structures from English needs further exploration.

\section*{Ethics Statement }\label{sec:ethical_consideration}
\paragraph{Copyright and Citation Issue} The copyright of individual datasets in \textsc{SummZoo} belongs to the original authors. The usage license of each dataset also applies to \textsc{SummZoo}. To ensure fair credit, when using \textsc{SummZoo} for evaluation, please also cite original papers, where individual datasets are introduced.

\paragraph{Data Availability and Safety} Pre-training and fine-tuning summarization data studied in this paper are mostly publicly available, otherwise we will provide links to the access application. Although filtering has been conducted in building the original datasets, some contents can contain uncomfortable descriptions, e.g., news coverage of violent crimes and events.

\paragraph{Usage of Large PLM} The GPT-3.5 model is used to generate text (summaries) for input documents of summarization tasks.
The generated text is only used for experiments and analysis, which are presented in corresponding sections.
No further usage, e.g., generating content for manuscripts, of GPT-3.5 or its family, is included in this paper.

\paragraph{Human Evaluation} We conduct human evaluation with the help of one judge, who obtained their postgraduate degree in the United Kingdom and has a solid experience in evaluating summarization tasks.
They were compensated through a payment of around $400$ USD for $450$ instances (\S~\ref{sec:human_evaluation}).

\section*{Acknowledgements}
We appreciate all reviewers and chairs from ACL 2023 for their valuable suggestions.
We thank Dan Iter, Hiteshi Sharma, Zicheng Liu, Sen Yang and Leyang Cui for their proofreading and inspiring discussion.


\newpage
\bibliography{anthology,custom}
\bibliographystyle{acl_natbib}

\appendix

\clearpage

\section{Datasets in SummZoo}\label{appendix:datasets_in_summzoo}

The final SummZoo contains following sub-tasks:

\paragraph{MultiNews~\cite{fabbri-etal-2019-multi}} is a large-scale multi-document summarization dataset. The task is to generate a summary given multiple news articles.

\paragraph{XSum~\cite{narayan-etal-2018-dont}} is an extreme text summarization dataset. Given a news article, the task is to generate a one-sentence summary.

\paragraph{Reddit-TIFU~\cite{kim-etal-2019-abstractive}} is a social post summarization dataset. The task is to generate a short summary for posts from the online discussion forum Reddit.\footnote{We categorize it into single document summarization task because the posts of each input are from the same user, centring one event.}
Compared with news text, the text in Reddit-TIFU is less formal and structured.

\paragraph{ArXiv~\cite{cohan-etal-2018-discourse}} is a long scientific paper summarization dataset collected from ArXiv, including articles of multiple domains, such as physics, computer science, etc.


\paragraph{WikiHow~\cite{koupaee2018wikihow}} is a large-scale instruction summarization dataset. The task is to generate a short summary given the multiple-step instruction.

\paragraph{SAMSum~\cite{gliwa-etal-2019-samsum}} is a written conversation summarization dataset for Messenger-style chit-chats. Both dialogue and summary are annotated by experts.

\paragraph{\textsc{DialogSum}~\cite{chen-etal-2021-dialogsum}} is a real-life scenario dialogue summarization dataset that covers a wide range of daily life dialogues, including diverse task-oriented dialogues. The testset of \textsc{DialogSum} provides three reference summaries for each dialogue, we report the averaged results.

\paragraph{QMSum~\cite{zhong-etal-2021-qmsum}} is a query-based meeting summarization dataset that is derived from Augmented Multi-party Interaction (AMI) corpus~\cite{kraaij2005ami},  the International Computer Science Institute (ICSI)~\cite{shribergicsi} and Committee Meetings.
The task is to generate a summary given a meeting and a query.

\section{Multi-Task Pre-Training Datasets}\label{appendix:multi-task_pre-training_datasets}

We use the following datasets for multi-task pre-training:


\paragraph{CNNDM~\cite{CNNDM}} is a large news summarization dataset that contains articles and paired human annotated summaries from CNN and Daily Mail.

\paragraph{BillSum~\cite{kornilova-eidelman-2019-billsum}} consists of the US Congressional and California state bills, and summaries written by Legislative Counsel.

\paragraph{PubMed~\cite{cohan-etal-2018-discourse}} contains large long scientific articles and human labeled abstracts.
Compared with ArXiv, which contains data from multiple domains, PubMed dataset focuses on the biomedical field.

\paragraph{GovReport~\cite{huang-etal-2021-efficient}} consists of long reports and summaries from government research agencies.

\paragraph{MediaSum~\cite{zhu-etal-2021-mediasum}} is an interview summarization dataset that contains $463.6k$ transcripts and summaries from NPR and CNN.

\paragraph{SummScreen~\cite{chen-etal-2022-summscreen}} consists of long TV series transcripts and human written recaps.

\paragraph{XWikis~\cite{perez-beltrachini-lapata-2021-models}} is a cross-lingual summarization dataset that contains Wikipedia articles and leading paragraphs in multiple languages.
We only use the English data that have paired documents and summaries.

To balance the training data size of different datasets, we perform down-sampling on over-sized datasets and up-sampling on low-resource datasets respectively. The statistics of resulting data for pre-training are shown in Table~\ref{tab:training_data}.

\begin{table}[t]
    \centering
    \renewcommand{\arraystretch}{1.1}
    \small
    \begin{tabular}{l|r|r}
    \thickhline
         \textbf{Dataset}  & \textbf{Raw Size}  & \textbf{Sam. Size}  \\
         \thickhline
         CNNDM & $287,227$   &  $287,227$\\
         BillSum & $23,455$ &  $113,694$\\
         PubMed  & $119,924$ &  $119,924$\\
         GovReport & $19,466$  &  $105,114$\\
         MediaSum  &$463,596$  &  $100,000$ \\
         SummScreen  &  $22,588$ &  \ \ $67,764$\\
         XWikis  &  $280,000$ &  $100,000$\\
         \hline
         Total  & -- &\ $893,723$

 \\
         \thickhline
    \end{tabular}
    \caption{Statistics of multi-task pre-training data. We combine 7 summarization tasks.
    Raw size is the number of input and output pairs of raw datasets. Sam. (Sampled) size is the size of balanced data that are actually used in multi-task pre-training.}
    \label{tab:training_data}
\end{table}

\begin{table*}[t]
\renewcommand{\arraystretch}{1.1}

    \centering
    \small
    \begin{tabular}{lc|ccc|ccc|ccc|ccc}
    \thickhline
        \multicolumn{2}{c|}{\multirow{2}{*}{\textbf{Task}}} &  
        \multicolumn{3}{c|}{\textbf{PEGASUS}} &
         \multicolumn{3}{c|}{\textbf{BART-PT}} &
        \multicolumn{3}{c|}{\textbf{MultiBART}}  &   
        \multicolumn{3}{c}{\textbf{\textsc{UniSumm}}}  
        \\
          &  & $D_{R1}$  &  $D_{R2}$ &  $D_{RL}$ &  $D_{R1}$  &  $D_{R2}$&  $D_{RL}$&  $D_{R1}$  &  $D_{R2}$  &  $D_{RL}$ &  $D_{R1}$ &  $D_{R2}$  &  $D_{RL}$  \\
        \thickhline
        \multirow{2}{*}{MultiNews}  &\ \ 10
        & 0.37 & 0.30 & 0.21
        &  1.04 & 0.37 & 0.23
        & 0.68 & 0.20 & 0.10
        &  0.33 & 0.27 & 0.23 \\
        &  100  
        &  0.20 & 0.24 &0.22  
        & 0.11 & 0.23 & 0.21
        & 0.26 & 0.21 & 0.19
        &  0.19 & 0.30 & 0.29 \\
        \hline
        \multirow{2}{*}{XSum}  &\ \ 10 
        & 1.45 & 0.93 & 1.26
        & 1.60 & 0.54 & 1.05
        & 1.65 & 0.72 & 1.28
        & 1.21 & 0.78 & 1.15   \\
        &  100  
        & 0.37 & 0.31 & 0.31
        & 0.27 & 0.28 & 0.30
        & 0.11 & 0.08  &  0.05
        & 0.27 & 0.18 & 0.23
        \\
        \hline
        \multirow{2}{*}{Arxiv}  &\ \ 10  
        &  0.57 & 0.09 & 0.28
        & 1.08 & 0.54 & 0.87
        &  0.32 & 0.36 &0.29
        &0.93 & 0.31 &  0.83  \\
         &  100  
         & 0.55 & 0.17 & 0.35
         & 0.83 & 0.38 & 0.76
         & 0.64 & 0.19 & 0.60
         & 0.54 & 0.18 & 0.54
         \\
        \hline
        \multirow{2}{*}{WikiHow}  
        &\ \ 10  
        & 0.79 & 0.25 &0.42
        & 0.66 & 0.35 & 0.56
        & 0.66 & 0.46 & 0.48
        & 0.40 & 0.31 & 0.48

        \\
        &  100  
        & 0.46 & 0.21 & 0.31
        & 0.25 & 0.15 &0.22
        & 0.38 & 0.26 & 0.31
        & 0.21 & 0.10 & 0.15
        \\
        \hline
        \multirow{2}{*}{Reddit}   &\ \ 10  
        &  0.83 & 0.28 & 0.76
        & 1.61 & 0.57 & 1.00
        & 1.20 & 0.49 & 0.78 
        &1.16 &  0.64 & 1.01\\
          &  100 
         &  0.71 & 0.31 & 0.50
         &	0.72 & 0.39 & 0.57
         & 0.68 & 0.43 & 0.61
         & 0.52 & 0.26 & 0.49
         \\
        \hline
        \multirow{2}{*}{\textsc{DialogSum}}  &\ \ 10  
        & 1.18 & 0.90 & 1.13
        & 0.96 & 0.68 & 0.65 
        & 1.46 & 1.01 & 1.02
        & 0.99 &0.76 & 0.80
        \\
        &  100  
        &  0.83 & 1.01 & 0.85
        & 0.90 & 1.08 & 0.83
        & 1.01 & 1.16 & 0.95
        & 0.91 & 1.10 & 1.00
        \\
        \hline
        \multirow{2}{*}{SAMSum}  &\ \ 10  
        & 1.61 & 1.19 &1.24
        & 1.58 & 1.44 & 1.19
        & 1.91 & 1.69 & 1.51
        & 1.07 & 0.82 & 0.83
        \\
        &  100  
        &0.47 & 0.39 & 0.60
        &0.29 & 0.54 & 0.57
        & 0.40 & 0.47 & 0.50
        & 0.47 &0.30 & 0.41
        \\
        \hline
        \multirow{2}{*}{QMSum}  &\ \ 10  
        & 0.84 & 0.52 & 0.60
        & 0.75 & 0.45 & 0.36
        & 0.71 & 0.42 & 0.20
        & 0.45 & 0.57 & 0.30
        \\
        &  100  
        &   0.72 & 0.82 & 0.72 
        &   0.55 & 0.55 & 0.41
        &	0.34 & 0.32 & 0.24 
        & 0.30 & 0.23 & 0.17
         \\
         \hline
        \multirow{2}{*}{Average}  &\ \ 10  
        & 0.96 & 0.56 & 0.74
         & 1.16 & 0.62 & 0.74
         &1.07 &0.69 & 0.71
         & 0.82 & 0.56 & 0.70
        \\
         &  100  
         & 0.54  & 0.43  & 0.48
         &0.49  & 0.45 & 0.48
         & 0.48 & 0.39 & 0.43
         & 0.43 & 0.33 & 0.41
         \\
        \thickhline
    \end{tabular}
    \caption{Comparison of model robustness towards different few-shot samples. We report the standard deviations of \textsc{Rouge} scores on $5$ sets of few-shot samples provided in \textsc{SummZoo} for each task and setting. 
    $D_{R1}$, $D_{R2}$ and $D_{RL}$ mean the standard deviations of \textsc{R1}, \textsc{R2} and \textsc{RL}, respectively. Lower standard deviation indicates the model is more robust towards different few-shot samples.
    The bottom block presents the averaged results of all 8 sub-tasks.
    }
    \label{tab:full_robustness}
\end{table*}

\begin{table}[htb]
    \setlength\tabcolsep{2.5pt}
    \renewcommand{\arraystretch}{1.1}
    \centering
    \small
    \begin{tabular}{lc|c|c|c}
    \thickhline
        \multicolumn{2}{c|}{\multirow{2}{*}{\textbf{Task}}} & $P_{0.01}$+$ L_{0.01}$  &$P_{0.05}$+$L_{0.05}$
         &  $P_{0.01}$+$L_{0.05}$
        \\
          &  &R2  &  R2  &  R2   \\
        \thickhline
        \multirow{2}{*}{MN}  &\ \ 10
        & 	15.32	&	15.00	&	15.19	 \\
        &  100  
        &	15.47	&	15.82	&	15.86		 \\
        \hline
        \multirow{2}{*}{XSum}  &\ \ 10 
        & \ \ 6.52	& \ \	6.41	&\ \	7.20		  \\
        &  100  
        &11.57 &	11.30	&	11.36	
        \\
        \hline
        \multirow{2}{*}{Arxiv}  &\ \ 10  
        & 	15.50 & 15.20& 	15.38		\\
         &  100  
         &	16.50	&	16.15	&	16.42		
         \\
        \hline
        \multirow{2}{*}{WH}  
        &\ \ 10  
        & \ \	9.48 & \ \ 	9.37	&\ \	9.35			
        \\
        &  100  
        &	11.81	&	11.72	&	11.73	
        \\
        \hline
        \multirow{2}{*}{Reddit}   &\ \ 10  
        & \ \ 5.72  &  \ \ 5.55 & \ \ 5.60  		 \\
          &  100 
         &\ \ 	6.17	&\ \	6.23	&\ \	6.17	\\
        \hline
        \multirow{2}{*}{\textsc{DS}}  &\ \ 10  
        &	13.39 &	13.26	& 13.38		 \\
        &  100  
        &	15.71	&	15.74	&	15.64			\\
        \hline
        \multirow{2}{*}{SS}  &\ \ 10  
        & 	18.55 	&	18.38 &	18.53 		 \\
        &  100  
        & 	20.93	 	&	20.96 	&	20.65	 
        \\
        \hline
        \multirow{2}{*}{QMSum}  &\ \ 10  
        & 	12.06	 	&	12.04	 &	12.12	 	 \\
        &  100  
        & 13.40	&	13.73	&	13.89\\
         \hline
        \multirow{2}{*}{Average}  &\ \ 10  
        &12.07 &	11.90	&	12.09	
        \\
         &  100  
         &	13.95 &	13.96	&	13.97		
         \\
        \thickhline
    \end{tabular}
    \caption{Results of \textsc{UniSumm} using different combinations of weight decay rates for multi-task training. $P_{0.01}$ indicates the weight decay rate for prefix parameters is $0.01$ and  $L_{0.01}$ indicates the weight decay rate for LM parameters is $0.01$.}
    \label{tab:decay_results}
\end{table}

\begin{table*}[t]
\renewcommand{\arraystretch}{1.5}
    \centering
    \small
    \begin{tabular}{l|p{13.5cm}}
    \thickhline
    \multicolumn{2}{c}{\textbf{MultiNews}}\\
    \thickhline
        Input text  & paris (ap)  the pompidou centre in paris hopes to display a long-vanished picasso painting in may, now that it has been recovered by u.s. customs authorities. this undated photo provided by the united states department of justice, shows a cubist painting entitled the hairdresserby pablo picasso. authorities say the painting worth millions of dollars was ... ||||| a stolen picasso worth millions of dollars was shipped to the u.s. in a package that described it as a \$37  "art craft "  but it will soon be on its way back to france...    \\
        \hline
        PEGASUS & – a 1911 cubist painting worth millions of dollars has been recovered after being mysteriously transported from france to the us nearly a decade ago. the pablo picasso painting, titled la coiffeuse, was reported missing from a pompidou storeroom in paris in 2001 \underline{and was last seen in a package shipped to the us from belgium}, the ap reports. federal prosecutors in brooklyn filed papers thursday to seize the painting, which was hidden in a climate-controlled storage facility in queens, ny, and will soon be returned to france, reports the new york times. the center for art recovery in paris says it hopes to have the painting back in may, at a time when the cultural world is reeling from islamic state videos showing the destruction of statues in iraq. "this is a true comfort for us at a time when the cultural world is reeling from the islamic state video showing the destruction of statues in iraq," says alain seban, the director of the pompidou centre in paris. "it is a work of art that represents the genius of pablo picasso and was bequeathed to us in 2001 by the master of france's national museum."\\
        \hline
        GPT-3.5 & – the new york times has a front-page story friday detailing how former fbi chief james comey tried to instruct president trump about the proper boundaries that should exist between them.\\
        \hline
        {\makecell[l]{\textsc{UniSumm}}} & - a century-old picasso painting missing from paris for more than a decade resurfaced in the united states under false pretenses as a \$37 holiday-themed "art craft," the ap reports. the 1911 painting, called "la coiffeuse," \underline{was unearthed in december in a fedex shipment from belgium to newark}. the canvas had been smuggled out of a storeroom of the centre georges pompidou, the paris museum and arts center. officials there thought it was safe and sound until they received a loan request for it in 2001. they searched the storerooms and could not find it. on dec. 17, someone going by "robert" with an address in belgium brought the painting to a climate-controlled warehouse in long island city and was seized, the ap reports. the us attorney for the eastern district of new york filed a civil complaint thursday to forfeit the picasso and return it to france. the canvas, titled "la coiffeuse," was bequeathed to the national museums of france in 1998. the pompidou director alain seban said in a statement friday that he hopes the work can be exhibited again publicly in may. no arrests have been made in the case.  \\
        \hline
        {Gold} & - a picasso painting that was found to have vanished from a paris museum more than a decade ago has \underline{turned up 2014 in the us, in a package shipped from belgium.} its papers identified it as a \$37 "art craft/toy" and also included the line "joyeux noel," or merry christmas, the new york times reports. a man named "robert" attempted to send the package to a climate-controlled warehouse in queens, new york, in december, but custom officials at the port of newark seized what turned out to be picasso's 1911 la coiffeuse (the hairdresser). french museum officials traveled to new york last month and confirmed the find is indeed the missing picasso work, which the centre georges pompidou realized was missing from its storerooms in 2001 following a loan request; it was then valued at more than \$2.5 million. court documents don't specify whether the sender or would-be recipient have been identified, nbc news reports, but "a lost treasure has been found," us attorney loretta lynch said in a statement. she filed a civil complaint yesterday that will have the painting returned to france. the pompidou's director, who called the painting's rediscovery a "true comfort," hopes the painting can go on display at the museum as early as may, the ap reports. (another picasso work was stolen days before this one was found.) \\
    \thickhline
    \end{tabular}
    \caption{Case study (a): Gold and generated summaries by 1-shot \texttt{text-davinci-002} (GPT3.5), 100-shot PEGASUS and \textsc{UniSumm} on MultiNews, respectively.
    ``|||||'' seperates different reports in the input. }
    \label{tab:case_multinews}
\end{table*}

\begin{table*}[h]
\renewcommand{\arraystretch}{1.5}
    \centering
    \small
    \begin{tabular}{l|p{13.5cm}}
    \thickhline
    \multicolumn{2}{c}{\textbf{QMSum}}\\
    \thickhline
        Input text  &\textbf{what did barry hughes think about draft changes to cps guidance when talking about the divergence between the law in england and wales?} suzy davies am: thank you. i was going to ask this question later, but i think it fits in . i've forgotten what i was going to ask; this is terrible. oh no, that's right obviously, this act is on a particular trajectory at the moment, and it's due to be introduced before the end of this assembly. do you think it would be better for us as scrutinisers of this act if we could see the draft changes to cps guidance on the public interest test before we make our final decision? barry hughes: i honestly don't think that would necessarily be helpful. i've had some discussions with kwame, who would have an involvement in this. ... \\
        \hline
        {PEGASUS} &barry hughes thought it would be better for the members to see the draft changes to cps guidance on the public interest test before making their final decision. he added that they would need to work that up as they went along, and it would be a risk of putting the cart before the horse.\\
        \hline
        GPT-3.5 & Barry Hughes thinks that it would not be helpful for the public to see the draft changes to the CPS guidance on the public interest test before the final decision is made on the bill. He believes that it would be better to take the general principles set out in the code for crown prosecutors and provide more specific details around them that relate to the issues being discussed in the bill.\\
        \hline
        {\makecell[l]{\textsc{UniSumm}}} &barry hughes thought that the draft changes to cps guidance would not necessarily be helpful. he thought that they would simply want to take the present public interest factors set out very clearly in the code for crown prosecutors, and provide a degree of detail around those that relates more specifically to the issues that they were discussing. so, it would be taking principles of generality and according them a degree of specificity. and they'd need to work that up as they go along.\\
        \hline
        {Gold} &barry hughes thought draft changes to cps guidance would not necessarily be helpful to the scrutinisers, and simply wanted to take the present interest factors. barry thought they needed to work for the principles of generality and a degree of specificity. \\
    \thickhline
    \end{tabular}
    \caption{Case study (b): Gold and generated summaries by 1-shot \texttt{text-davinci-002} (GPT3.5), 100-shot PEGASUS and \textsc{UniSumm} on QMSum, respectively. The \textbf{blod} is the query. }
    \label{tab:case_qmsum}
\end{table*}

\section{Implementation Details}\label{appendix:Implementation_details}
We use BART-large~\cite{lewis-etal-2020-bart} to initialize the summarization model of \textsc{UniSumm}.
All experiments are conducted on NVIDIA A100 GPU with PyTorch 1.11.
The max input length and target length are set to 2,048 and 400.
The hyper-parameter choice is based on previous few-shot summarization work~\cite{DBLP:conf/icml/ZhangZSL20, fabbri-etal-2021-improving, DBLP:conf/aaai/ChenS21a} and empirical consideration.
For multi-task pre-training, we initialize from BART-large, and train the model on 16 GPUs with 300,000 steps, batch size of 32, learning rate of 1.5e-5, and warm-up with 4,000 steps.
For few-shot tuning, we prefix-tune the model on 4 GPUs with 100 and 1000 steps for 10-shot and 100-shot, respectively, with batch size of 32, learning rate of 1.5e-4, and warm-up with 10\% of the training steps.
For XSum, the training steps are set to 10 and 100 for 10-shot and 100-shot, respectively, while other configurations are unchanged.


\section{Model Robustness}\label{appendix:model_robustness}
Table~\ref{tab:full_robustness} shows the standard deviations of \textsc{Rouge-1}, \textsc{Rouge-2} and \textsc{Rouge-L} scores on 5 different sets of few-shot samples in \textsc{SummZoo}.
Overall, \textsc{UniSumm} shows the least standard deviations on most metrics across tasks in both settings, suggesting it is most robust and stable towards different selections of training samples.

\section{Human Evaluation}\label{appendix:human_evaluation}
Following~\citet{kryscinski-etal-2019-neural,kryscinski-etal-2020-evaluating}, we conduct human evaluation from 4 dimensions, which can offer a more robust and holistic perspective to understand summarization system~\cite{zhong-etal-2022-towards}:

\begin{itemize}
    \item \emph{Fluency} evaluates the quality of individually generated sentences, including grammar, word order, etc;
    \item \emph{Coherence} evaluates the collective quality of generated summaries;
    \item \emph{Relevance} evaluates the importance of information in the generated summaries; 
    \item \emph{Consistency} evaluates the factual alignment of the generated summary against the input document. 
\end{itemize}

We ask a judge to give scores from $1$ to $5$ along these 4 dimensions.
Higher score indicates better quality.
The judge is a postgraduate student, who studied in the United Kingdom and has solid experience in evaluating summarization tasks.

\section{Case Study}
\label{appendix:case_study}
We  qualitatively demonstrate the advantages of \textsc{UniSumm} (100-shot) using cases from MultiNews and QMSum, and present an error analysis using case from WikiHow.

As shown in Table~\ref{tab:case_multinews} (MultiNews), we see that the \textsc{UniSumm} generates a summary with similar events and faithful descriptions compared with the gold summary.
However, PEGASUS generated summary contains factual errors (``\emph{... was last seen in a package shipped to the us from belgium.}'') while the summary generated by \textsc{UniSumm} (``\emph{... unearthed ... shipment from belgium to newark}'') is consistent with the gold summary and input (``\emph{... turned up ... shipped from belgium.}'').
This shows that \textsc{UniSumm} has the ability to collect important information from multiple news reports and generate high-quality summaries, which is a task that the model has never seen during multi-task pre-training.

Also, as shown in Table~\ref{tab:case_qmsum} (QMSum), 
compared with gold summary, although the summary generated by \textsc{UniSumm} is longer, it is highly relevant to the query.
And \textsc{UniSumm} properly rephrases the key utterance from the source meeting into an objective description, which suits the characteristic of conversation summarization.
In contrast, the summary generated by \textsc{PEGASUS} misses important contents and contains irrelevant sentences compared with \textsc{UniSumm} and human annotation.
This evidence shows that \textsc{UniSumm} successfully learns important characters of query-based meeting summarization task with only 100 samples.

An error case where \textsc{UniSumm} fails can be found in Table~\ref{tab:case3} (WikiHow). 
UniSumm mistakenly generates ``\emph{...matches the text of the letter...}'', where the ground truth should be the ``\emph{...matches…the one (address)...on the envelope}''.
Moreoever, the summary generated by UniSumm is a bit repetitive in wording, e.g., serveral repeated phrases ``\emph{... on the inside of the letter...}''.

We present more cases in Table~\ref{tab:case2} (ArXiv and \textsc{DialogSum}), Table~\ref{tab:case3} (XSum) and Table~\ref{tab:case4} (SAMSum and Reddit).
Overall, we find that \textsc{UniSumm} is capable of generating very fluent, relevant, faithful and human-like summaries on diverse unseen tasks.
This verifies \textsc{UniSumm}'s great generalization ability in the few-shot scenario.

\section{Influence of Weight Decay}\label{appendix:influence_of_wedight_decay}

In \S~\ref{sec:weight_decay}, we design a separated weight decay strategy to circumvent negative transfer in multi-task learning. 
In Table~\ref{tab:decay_results}, we examine whether
the combination of different weight decay rates ($d_p$ for prefixes and  $d_l$ for the summarization model) is beneficial.
Specifically, we report \textsc{Rouge}-2 scores on \textsc{SummZoo} with different combinations of weight decay rates.
We can see that the model performs the best with $d_p=0.05$ and  $d_l=0.01$.
And this asymmetrical weight decay is especially helpful to 10-shot XSum, which is more distinct from pre-training summarization tasks and relies more on general summarization knowledge.

\begin{table*}[t]
\renewcommand{\arraystretch}{1.5}
    \centering
    \small
    \begin{tabular}{l|p{13.5cm}}
    \thickhline
    \multicolumn{2}{c}{\textbf{ArXiv}}\\
    \thickhline
        Input text  &  rapid progress in the design and manufacture of optical fiber systems is a result of worldwide demand for ultra - high bit - rate optical communications . this explains the growing interest of the soliton community in soliton - based optical fiber communication systems . this area of research was considerably advanced in recent years @ xcite . the most remarkable results include the application of the concept of the dispersion management to \_ temporal optical solitons \_ and soliton - based optical transmission systems , and the discovery of the so - called \_ dispersion managed soliton \_ . high - speed optical communications require effective components such as high - performance broadband computer networks that can be developed by employing the concept of the bit - parallel - wavelength ( bpw ) pulse transmission that offers many of the advantages of both parallel fiber ribbon cable and conventional wavelength - division   ... \\
        \hline
        PEGASUS & the physical properties of multi - frequency solitary waves propagating in optical fibers and waveguides are discussed in the context of the soliton - based optical dispersion management. the emphasis is placed on the role of the spatial dispersion in the propagation of such solitary waves. it is shown that the propagation of such waves in optical fibers and waveguides is affected not only by the parametric difference between the pulse and the beam velocity, but also by the spatial variation of the dispersion profile. as a consequence, we argue that the role of the spatial dispersion in the propagation of solitary waves in optical fibers and waveguides is not limited to the parametric difference between the pulse and the beam velocity, but also includes the role of the spatial dispersion itself.\\
        \hline
        GPT-3.5 & This article discusses the various types of solitary waves that can occur in nonlinear optical systems, with a focus on how to extract information about the propagation characteristics from experimental data. The examples of temporal and spatial solitons are given, along with a discussion of the differences between them. The difficulties associated with experimental measurements are also addressed, along with suggestions for ways to overcome them.\\
        \hline
        {\makecell[l]{\textsc{UniSumm}}} &  the aim of the present paper is to present a versatile scheme for the study of spatial and temporal solitary waves . this approach is based on the theory of spatio - temporal wave propagation , where the spatial beam profile along the transverse direction is similar to the temporal pulse profile in fibers . in all such systems , the nonlinear interaction between the waves of two ( or more ) frequencies is the major physical effect that can support coupled - mode multi - frequency solitary waves . we will illustrate three common examples : multi - wavelength solitary waves in bit - parallel - wavelength optical fiber links , multi - colour spatial solitons due to multistep cascading in optical waveguides with quadratic nonlinearities , and quasiperiodic solitons in the fibonacci superlattices .\\
        \hline
        {Gold} & we discuss several novel types of multi - component ( temporal and spatial ) envelope solitary waves that appear in fiber and waveguide nonlinear optics . in particular , we describe multi - channel solitary waves in bit - parallel - wavelength fiber transmission systems for high performance computer networks , multi - colour parametric spatial solitary waves due to cascaded nonlinearities of quadratic materials , and quasiperiodic envelope solitons due to quasi - phase - matching in fibonacci optical superlattices . 2 \\
        
    \thickhline
    \multicolumn{2}{c}{\textsc{\textbf{DialogSum}}}\\
    \thickhline
        Input text  & person1: ms. dawson, i need you to take a dictation for me. person2: yes, sir... person1: this should go out as an intra-office memorandum to all employees by this afternoon. are you ready? person2: yes, sir. go ahead. person1: attention all staff... effective immediately, all office communications are restricted to email correspondence and official memos. the use of instant message programs by employees during working hours is strictly prohibited. ...  \\
        \hline
        PEGASUS & person1 advises dawson to sign an intra-office memorandum restricting office communications to email correspondence and official memos. person2 asks if this applies to internal and external communications. person1 tells them so.\\
        \hline
        GPT-3.5 & person1 needs Ms. Dawson to take a dictation for an intra-office memorandum that is to be distributed to all employees by 4 pm. The memo is to inform all staff that effective immediately, all office communications are restricted to email correspondence and official memos; the use of instant message programs by employees during working hours is strictly prohibited. At first offense, the employee will receive a warning and be placed on probation. The second offense will result in termination.\\
        \hline
        \textsc{UniSumm}&  person1 asks ms. dawson to take a dictation for person1. person1 restricts office communications to email correspondence and official memos and prohibits the use of instant message programs by employees during working hours. \\
        \hline
        {Gold} & ms. dawson helps person1 to write a memo to inform every employee that they have to change the communication method and should not use instant messaging anymore.\\
    \thickhline
    \end{tabular}
    \caption{Case study (c): Gold and generated summaries by 1-shot \texttt{text-davinci-002} (GPT3.5), 100-shot PEGASUS and \textsc{UniSumm} on ArXiv and \textsc{DialogSum}, respectively.}
    \label{tab:case2}
\end{table*}
\begin{table*}[t]
\renewcommand{\arraystretch}{1.5}
    \centering
    \small
    \begin{tabular}{l|p{13.5cm}}
    \thickhline
    \multicolumn{2}{c}{\textbf{WikiHow}}\\
    \thickhline
        Input text  &  before folding the letter, make sure that the address on the inside of the letter matches with the one that you have written or printed on the envelope.this will help avoid any mixups. don't forget to check that you've signed the letter. make sure that the text is facing up and towards you. this is your last chance to proofread and double check that you haven't forgotten anything.for example, have you dated the letter? are there any spelling or grammar mistakes? , take the bottom of the letter and fold it so that it is about half an inch (about 1 cm) from the top of the page.you can place the envelope underneath the letter to use as a guide. make sure that when you have folded the letter up, it is small enough to fit inside the envelope. make sure to line up all the outside edges of the paper before creasing so that you don't end up with a crooked fold. if the fold is crooked, your letter may not fit into the envelope.use a ruler to make the crease nice and sharp. you can do this by holding a ruler sideways and using the thin edge. slide the edge of the ruler along the letter on the creases so they are nice and flat. you will now take the right part of the letter and fold it about one third of the way in.line up the top and bottom edges of the letter and then crease the fold. take the other side of the letter and fold it one third of the way in. this will be the same thing that you did with the right side.line up the top and bottom of the edges so they are perfectly straight before creasing. the last crease you made should go into the bottom of the envelope first. place it so that the folds face the backside of the envelope.this will make it easy for your recipient to find where to begin unfolding the letter.  \\
        \hline
        PEGASUS & take the top of the letter and fold it so that it is about half an inch (about 1 cm) from the top of the page. place the envelope underneath the letter. make sure that the envelope is flat on the inside. crease the folded side of the letter.\\
        \hline
        GPT-3.5 & The document explains how to fold a letter so that it will fit into an envelope. First, make sure that the address on the letter matches the address on the envelope. Next, fold the bottom of the letter up to the top of the page, making sure to crease the fold. Then, fold the right side of the letter in to the middle, and repeat with the left side. Finally, insert the letter into the envelope so that the folds are facing the back.\\
        \hline
        \textsc{UniSumm} &   check that the address on the inside of the letter matches the text of the letter to make sure that the address on the inside of the letter matches with the one on the page. make sure that the text matches with the one on the inside of the letter. make sure that the envelope is small enough to fit into the envelope. 
        \\
        \hline
        {Gold} & check the address. lay the letter on a table. fold the bottom half up. crease the fold. fold the right half of the letter inwards. fold the left half of the letter inwards. turn the letter sideways and insert into the envelope. \\
    \thickhline
    \multicolumn{2}{c}{\textbf{XSum}}\\
    \thickhline
        Input text  & the sunday times says the missile veered off course during a test in june last year - weeks before the commons voted to spend 40bn renewing trident . questioned by andrew marr, the pm refused to say four times if she had known about the test ahead of the vote . the snp's nicola sturgeon called for a ' full disclosure ' of what happened . according to the sunday times, an unarmed trident ii d5 missile veered off in the wrong direction towards the us - instead of towards africa - when it was launched from a british submarine off the coast of florida . in july - days after mrs may had become prime minister - mps voted overwhelmingly in favour of replacing trident . during the debate, mrs may told mps it would be ' an act of gross irresponsibility ' for the uk to abandon its nuclear weapons . mps backed its renewal by 472 votes to 117 . however, all 52 snp mps voted against it - as did labour leader jeremy corbyn . when asked on the bbc's andrew marr show whether she had known then that a misfire had happened, mrs may said: ' i have absolute faith in our trident missiles . ' when i made that speech in the house of commons, what we were talking about was whether or not we should renew our trident . ' she was asked a further three times - but did not answer the questions . the ministry of defence did not give details of the test process but said it had been a success . scottish first minister, mrs sturgeon - a long-standing opponent of trident, whose submarines are based at faslane, on the river clyde - said the apparent misfire was a ' hugely serious issue ' . she tweeted: ' there should be full disclosure of what happened, who knew what/when, and why the house of commons wasn't told . ' meanwhile, mr corbyn said the reports called for ' a serious discussion ' . ...\\
        \hline
        PEGASUS & the prime minister has been accused of " side-stepping " questions about a submarine-launched nuclear-capable missile which misfired during a test.\\
        \hline
        GPT-3.5 & trident is the uk's nuclear weapons system, consisting of submarines, missiles and warheads. it is operated by the royal navy and has been in use since the 1990s. reports that a missile test in june 2016 went wrong have raised questions about the system's effectiveness and whether it was deliberately kept quiet ahead of a key parliamentary vote on its renewal.\\
        \hline
        \textsc{UniSumm} &  a nuclear missile launched a british submarine failed during a test, according to a report in the sunday times. \\
         \hline
        {Gold} & theresa may is coming under pressure to say whether she knew about a reported misfire of the uk's nuclear weapons system before a crucial commons vote. \\
    \thickhline
    \end{tabular}
    \caption{Case study (d): Gold and generated summaries by 1-shot \texttt{text-davinci-002} (GPT3.5), 100-shot PEGASUS and \textsc{UniSumm} on WikiHow and XSum, respectively.}
    \label{tab:case3}
\end{table*}

\begin{table*}[t]
\renewcommand{\arraystretch}{1.5}
    \centering
    \small
    \begin{tabular}{l|p{13.5cm}}
    \thickhline
    \multicolumn{2}{c}{\textbf{SAMSum}}\\
    \thickhline
    Input text & cora: have you heard how much fuss british media made about meet and greet with james charles in birmingham? ellie: no...! what happened? cora: well, there was a meet and greet with james charles in one of the malls in birmingham and about 8000 fans showed up for it. cora: it cause a gridlock around the mall and - of course - british media had to make some (quite negative) comments on it. ellie: they came for sister james?! >:( ellie: i sister snapped!! :p :d cora: haha :d cora: you shouldn't watch so much youtube, you're getting weirder and weirder. :d ellie: sister shut up :p so, what did they say? cora: ;) : * daily mail" was surprised that a meet and greet with a "virtually unknown" youtuber gathered 8000 people. :p cora: a host from lbc tried to find an answer to an unanswerable question: "who is james charles?" ... \\
    \hline
    PEGASUS & ellie's meet and greet with james charles in birmingham got some negative comments from british media. cora thinks it's ridiculous that a man can wear makeup.\\
    \hline
    GPT-3.5 & cora and ellie were discussing the meet and greet with james charles in birmingham. 8000 fans showed up and it caused a gridlock. british media was surprised and some hosts couldn't wrap their heads around the fact that a guy can wear makeup.\\
    \hline
    \textsc{UniSumm} &  about 8000 fans showed up for the meet and greet with james charles in birmingham. \\
    \hline
    Gold & there was a meet-and-greet with james charles in birmingham which gathered 8000 people. \\
    \thickhline
    \multicolumn{2}{c}{\textbf{Reddit}}\\
    \thickhline
    Input Text & tifu making lunch. made myself a lush sandwich for lunch today. smoked ham, cheddar cheese, a dash of bbq sauce on thick cut wholemeal bread. popped it in a small tupperware box, for freshness. i found some bn type biscuits in the cupboard, so i pop them in the box on top of my sandwiches. think nothing of it. eventually lunch time comes around. i discover, to my horror, that the biscuits have absorbed the moisture from the sandwich. and now i have a dried out sandwich and soggy biscuits\\
            \hline

    PEGASUS & made a luch sandwich for lunch, only to have it ruined by soggy biscuits. i'm completely fucked.\\
            \hline
    GPT-3.5 & made a sandwich and it turned out terrible because the biscuits absorbed all the moisture.\\
    \hline
    \textsc{UniSumm} & made a big sandwich, with cheddar cheese, bbq sauce and bn type biscuits. popped them in a tupperware box, and discover they've dried out\\
            \hline

    Gold & don't put your sandwich and biscuits in same box, unless you like dry bread and soggy biscuits.\\
    \thickhline
    \end{tabular}
    \caption{Case study (e): Gold and generated summaries by 1-shot \texttt{text-davinci-002} (GPT3.5), 100-shot PEGASUS and \textsc{UniSumm} on SAMSum and Reddit, respectively.}
    \label{tab:case4}
\end{table*}


\end{document}